%% file: main.tex
\documentclass{article}

\usepackage[square,numbers]{natbib}
\usepackage[preprint]{neurips_distshift_2021}

\usepackage[utf8]{inputenc} %
\usepackage[T1]{fontenc}    %
\usepackage[colorlinks=true]{hyperref}       %
\usepackage{url}            %
\usepackage{booktabs}       %
\usepackage{amsfonts}       %
\usepackage{nicefrac}       %
\usepackage{microtype}      %
\usepackage{xcolor}         %
\usepackage{pgfplots}
\pgfplotsset{compat=1.17}
\usepackage{tikz}
\usepackage[caption=false]{subfig}
\usepackage{comment}
\usepackage{nicefrac}
\usepackage{amsmath}
\usepackage{multirow}
\usepackage{colortbl}
\usepackage{graphicx}
\usepackage{bmpsize}

\definecolor{unet}{rgb}{1.0, 0.65, 0.0}
\definecolor{resnet}{rgb}{0.89, 0.82, 0.04}
\definecolor{densenet}{RGB}{128,0, 128}
\definecolor{pnas}{rgb}{0,0.72,0.92}
\definecolor{resnet_skip}{rgb}{1.0,0.0,0.56}
\definecolor{drak_light_blue}{RGB}{1, 176, 239}
\definecolor{drak_dark_blue}{RGB}{51, 134, 155}
\definecolor{drak_red}{RGB}{254, 76, 82}
\definecolor{drak_green}{RGB}{25, 204, 153}
\definecolor{drak_orange}{RGB}{254, 151, 22}
\definecolor{drak_magenta}{RGB}{203, 154, 254}
\definecolor{drak_third}{rgb}{1, 1, 0.85}
\definecolor{drak_second}{rgb}{1, 0.8, 0.6}
\definecolor{drak_worst}{rgb}{1, 0.6, 0.6}

\newcommand{\worst}[1]{\textbf{#1} \cellcolor{drak_worst}}
\newcommand{\second}[1]{#1\cellcolor{drak_second}}
\newcommand{\third}[1]{#1 \cellcolor{drak_third}}

\title{A benchmark with decomposed distribution shifts for 360 monocular depth estimation}

\author{
  Georgios Albanis$^1$\thanks{Denotes equal contribution.} \quad Nikolaos Zioulis$^{1, 2}$\footnotemark[1] \quad Petros Drakoulis$^1$\footnotemark[1] \quad Federico Alvarez$^2$ \\
  \textbf{Dimitrios Zarpalas}$^1$ \quad \textbf{Petros Daras}$^1$
  \vspace{0.05in} \\
   $^1$ Centre for Research and Technology Hellas, Thessaloniki, Greece\\
   $^2$ Universidad Polit\'{e}cnica de Madrid, Madrid, Spain\\
   \texttt{\{galbanis, nzioulis, petros.drakoulis, zarpalas, daras\}@iti.gr}\\
   \texttt{fag@gatv.ssr.upm.es}\\
   \vspace{0.01in} \\
   \normalsize\href{https://vcl3d.github.io/Pano3D/}{vcl3d.github.io/Pano3D}
}

\begin{document}

\maketitle

\begin{abstract}
    In this work we contribute a distribution shift benchmark for a computer vision task; monocular depth estimation. Our differentiation is the decomposition of the wider distribution shift of uncontrolled testing on in-the-wild data, to three distinct distribution shifts. Specifically, we generate data via synthesis and analyze them to produce covariate (color input), prior (depth output) and concept (their relationship) distribution shifts. We also synthesize combinations and show how each one is indeed a different challenge to address, as stacking them produces increased performance drops and cannot be addressed horizontally using standard approaches.
\end{abstract}

\section{Introduction}

Data-driven methods are conditioned on the data which are available during the model development but are to be applied on real world data.
Considering that the former data distribution is $P_s = \sigma(P_r)$, which is the result of applying a sampling function $\sigma(\cdot)$ to the real world distribution $P_r$.
Typically, $P_s$ is separated into different splits $P_{trn}$ and $P_{val}$, used to train the model and validate its behaviour respectively, with the latter process driving model selection.
A data distribution shift can be described as the condition where the joint distribution $P$ of inputs and outputs differs between the training and test stages \cite{quinonero2009dataset}. 

This is an actual problem that many practical applications face, affecting their overall performance, robustness and reliability. 
The phenomenon is more prominent in tasks where annotated data collection is difficult and has been generally addressed in the literature as the domain shift \cite{stacke2019closer} or the generalization of data-driven models \cite{dubey2021adaptive,zhang2021out}, or otherwise as out-of-distribution robustness \cite{andreassen2021evolution}.
More information about out-of-distribution (OOD) learning and generalization can be found in recent surveys \cite{geisa2021towards,shen2021towards,yang2021generalized}.

Up to now, most works approach this problem in its general setting via zero-shot cross-dataset transfer experiments that aim at assessing model performance under a general distribution shift, considering two different samplings $P_{s^1} = \sigma_1(P_r)$ and $P_{s^2} = \sigma_2(P_r)$, as seen in Figure~\ref{fig:typical_vs_cross}.
A recent benchmark \cite{koh2021wilds} provided simultaneously data for sub-population shift, a special case of distribution shifts, and a generic domain generalization shift across a number of datasets and tasks.

\input{Figures/typical_vs_cross}

In this work, we contribute a novel benchmark for distribution shift performance assessment, in the context of a computer vision task notorious for its complex data collection processes; monocular depth estimation.
The novelty of our benchmark lies in the decomposition of generalized shift into components, expressed separately or in combination, via targeted test splits.

\section{The Pano3D Dataset}

Our benchmark relies on two recent 3D building scan datasets, Matterport3D (M3D) \cite{chang2017matterport3d} and GibsonV2 (GV2) \cite{xia2020interactive}, using modern synthesis to produce high quality spherical panoramas coupled with depth maps. Sample images can be found in Figure~\ref{fig:datasets}.
Specifically, we use M3D as a traditional in-distribution model development dataset and GV2 as a zero-shot cross-dataset transfer, out-of-distribution benchmark dataset.

\input{Figures/datasets}

For M3D, we consider its standard partitioning into train $P_{trn}^{M3D}$, validation $P_{val}^{M3D}$ and test $P_{tst}^{M3D}$ splits.
The GV2 splits represent another sampling of the real world domain, or otherwise a zero-shot cross-dataset transfer experiment.
Nonetheless, GV2 itself is partitioned into different splits, the \textit{tiny} $P_{tiny}^{GV2}$, \textit{medium} $P_{med}^{GV2}$, \textit{full} $P_{full}^{GV2}$ and \textit{fullplus} $P_{fullplus}^{GV2}$ splits\footnote{For the remainder of the document we ignore the \textit{full} split, which is kept aside for future training purposes.}.
After synthesizing coupled color and depth panoramas for all splits of both datasets, we analyze them and observe that it is possible to decompose them into three core distribution shifts. More on the characterization and decomposition of distribution shift can be seen on \cite{hu2021understanding,federici2021information,wiles2021fine}:

A \textcolor{drak_magenta}{\textbf{covariate}} distribution shift represents a shift of the input domain, which in our case is the color image's domain. 
As we rely on a synthesis approach (\textit{i.e.}~raytracing) to generate our data, we are also in control of the camera color transfer function.
Consequently, we can generate a shifted input distribution $P_{cov}$ using the M3D test split $P_{tst}^{M3D}$, where only the color domain has been shifted.

After examining the different splits' statistics we also observe a \textcolor{drak_orange}{\textbf{prior}} probability distribution shift manifesting at the tiny, $P_{tiny}^{GV2}$, and medium, $P_{med}^{GV2}$, splits which corresponds to $P_{prior}$, meaning that the output depth distribution has shifted from the training one $P_{trn}^{M3D}$.
Yet, the input (color) distribution is similar as the color camera transfer function is the same, and the context is also preserved to residential scenes.

Finally, analysing the \textit{fullplus} split, we observe a \textcolor{drak_green}{\textbf{concept}} distribution shift, which is the shifted context of the depicted scenes.
While Matterport3D (\textit{i.e.}~$P_{trn}^{M3D}$) only contains indoor residential scenes, the \textit{fullplus} split $P_{fullplus}^{GV2}$ presents varying scenes like supermarkets, garages, under construction buildings, etc., corresponding to $P_{conc}$.
At the same time though, the input (color) and output (depth) distributions are preserved between $P_{trn}^{M3D}$ and $P_{conc}$.

Notably, our benchmark decomposes the wider domain shift into three distinct distribution shifts.
But since we rely on synthesis processes, it is straightforward to combine distribution shifts, producing $P_{prior}^{cov}$ and $P_{conc}^{cov}$ by re-rendering the corresponding splits with a shifted color transfer function, essentially adding a covariant shift to the prior and concept ones.
This provides two extra combined distribution shift splits, with only the simultaneous prior and concept shifts missing.

Details can be seen in Figure~\ref{fig:disentangled}.
All of our data are publicly available at: \href{https://vcl3d.github.io/Pano3D/}{vcl3d.github.io/Pano3D}.

\input{Figures/disentangled}

\section{Analysis}
We support our benchmark with a set of zero-shot cross-dataset transfer experiments across the different distribution shifts.
We use a standard UNet \cite{ronneberger2015u} architecture training a supervised model with a complex objective similar to \cite{albanis2021pano3d}:

\begin{equation}
    \mathcal{L} = \lambda_{L1} \mathcal{L}_{L1} +  \lambda_{cos} \mathcal{L}_{cos} + \lambda_{grad} \mathcal{L}_{grad} + \lambda_{vnl} \mathcal{L}_{vnl},
\end{equation}

where $\mathcal{L}_{L1}$ is an L1 loss, $\mathcal{L}_{cos}$ is an angular loss defined on the surface normals, $\mathcal{L}_{grad}$ is the multi-scale gradient matching loss from \cite{ranftl2020towards}, and $\mathcal{L}_{vnl}$ is the virtual normal loss \cite{yin2019enforcing}.
All the independent term weights are equally weighted, \textit{i.e.}~$\lambda_i = 1.0 \, \forall \, i$
We initialize our model using \cite{he2015delving} and optimize it using a batch size of $4$ and the Adam optimizer \cite{kingma2014adam}, using a learning rate of $0.0002$ and its default momentum parameter values.

When training we only use $P_{trn}^{M3D}$ and for all experiments we calculate standard metrics for depth estimation \cite{eigen2014depth}, as well as boundary \cite{koch2018evaluation,hu2019revisiting} and normals RMSE and accuracies \cite{wang2020vplnet}.
We aggregate performance across the different traits (direct depth, boundary and smoothness) them using a set of indicators:
\begin{align}
\label{eq:indicators}
    i_d &= ((1 - \delta_{1.25}) RMSE)^{-1},\\
    i_b &= ((1 - \frac{(F_{0.25} + F_{0.5} + F_{1.0})}{3}) dbe^{acc})^{-1},\\
    i_s &= ((1 - \frac{(\alpha_{11.25^o} + \alpha_{22.5^o} + \alpha_{30^o})}{3}) RMSE^o)^{-1},
\end{align}
where $RMSE$ and $RMSE^o$ are the depth and normal angular errors respectively, $dbe^{acc}$ is the accuracy boundary error from \cite{koch2018evaluation}, $F_t$ is the F1-score for different edge thresholds from \cite{hu2019revisiting}, and $\delta_{d}$ and $\alpha_{a}$ are the accuracy under thresholds for the depth and surface normals from \cite{eigen2014depth} and \cite{wang2020vplnet} respectively.

Through these indicators we present an holistic view of how task performance is affected from the different distribution shifts. 
In the following subsections, we examine isolated distribution shifts as well as some of their combinations.

\label{ana}
\textbf{Decomposed Shifts: Varying the input, output  and combined domains.}
After training a supervised model on M3D's train split, we examine its performance on the different distribution shifts we have generated compared to that of the in-distribution test set. 
Figure~\ref{fig:indicators_all} illustrates the results using the indicators from Eq.~\eqref{eq:indicators}.
We observe a performance drop for all distribution shifts, with the covariate (\textcolor{resnet_skip}{magenta} box) and prior (\textcolor{resnet}{orange} box) being at about the same level, while the concept shift (\textcolor{pnas}{cyan}~box) presents the largest performance loss.
At the same time, combining two distribution shifts hurts performance even more, as shown by the combined distribution shifts (\textcolor{densenet}{violet} box).

\input{Figures/indicators_all}

\textbf{Photometric augmentations for covariate shift.}
Next, we examine the effect of training with photometric augmentations (\textit{i.e.}~brightness, contrast, hue shifts, and gamma corrections) and testing on the different (combined or not) distribution shifts.
Figure~\ref{fig:indicators_augm} presents the results comparing training with and without augmentations.
It is generally acknowledged that photometric augmentations address camera domain or color transfer function shifts, and our experiments verify this, as performance gains are only observed in the splits where covariate shifts manifests.

\input{Figures/indicators_augm}

\textbf{Pretraining for generalization boost.}
Another common assumption is that pretraining on large image datasets like ImageNet \cite{deng2009imagenet} helps address domain shifts. 
We perform another experiment, this time using the PNAS model \cite{liu2018progressive} with all hyperparameters preserved, and train one model initialized with pretrained weights and another one initialized using \cite{he2015delving}.
Figure~\ref{fig:indicators_pnas} presents the results when tested on our benchmark's different shifts.
Interestingly, we observe a performance boost in the splits where only a single distribution shift is present, where in contrast, the ones with two stacked distribution shifts show minimal gains.
This indicates that pretraining does not necessarily improve generalization -- in the form of more transferable initial features -- but, instead, only provides a better parameter initialization leading to higher quality parameters' optimization.

\input{Figures/indicators_pnas}

The full array of the conducted experiments and their detailed results can be found in Table~\ref{tab:Table1_supp}. 

\section{Conclusion}

Distribution shifting is pivotal to the real-world application of data-driven methods.
In this work, we contribute a distribution shift benchmark for an ill-posed dense computer vision task, with notoriously difficult data collection process.
Seeking to facilitate future research towards addressing this challenging problem, we decompose distribution shift to input (covariate), output (prior) and their relationship (concept), providing an experimental baseline for further experimentation and understanding.

\input{Tables/Table1_supp}

\begin{ack}
This work was supported by the EC funded H2020 project ATLANTIS [GA 951900].

\end{ack}

\medskip

\bibliographystyle{unsrt}
\bibliography{bibliography}

\end{document}

%% file: Figures/typical_vs_cross.tex
\begin{figure*}[!htbp]
\centering
\subfloat{\includegraphics[scale=0.54]{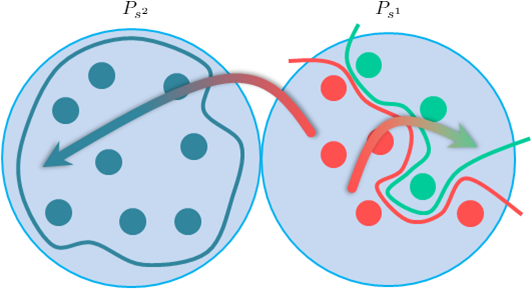}}
\caption{Typical training setting vs. zero-shot cross-dataset transfer. Each \textcolor{drak_light_blue}{light blue} disk represents a distinct sampling of $R$, a dataset. \textcolor{drak_red}{Red}/\textcolor{drak_green}{green} represent the typical train/test splits and \textcolor{drak_dark_blue}{dark blue} a zero-shot cross-dataset transfer test split. For simplicity, we omit the validation splits, considering them part of the train splits.}
\label{fig:typical_vs_cross}
\end{figure*}

%% file: Figures/datasets.tex
\begin{figure*}[!htbp]
\centering
\subfloat{\includegraphics[width=0.99\linewidth]{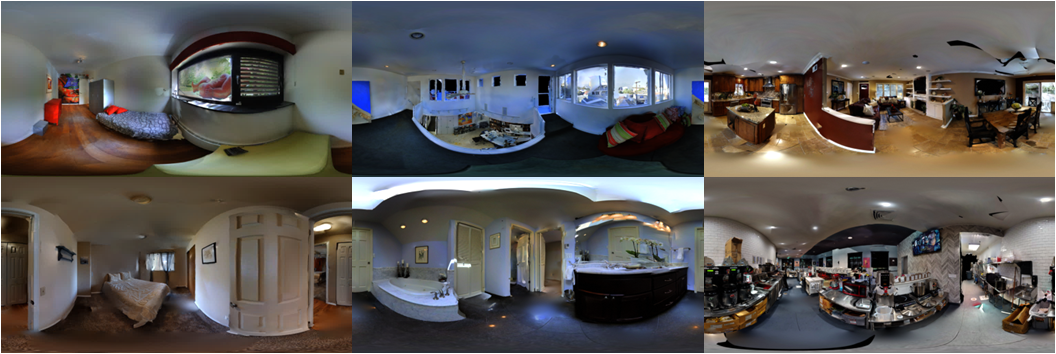}}
\caption{Rendered panoramic images of Matterport3D (top) and GibsonV2 (bottom).}
\label{fig:datasets}
\end{figure*}

%% file: Figures/disentangled.tex
\begin{figure*}[!htbp]
\centering
\subfloat{\includegraphics[width=0.99\linewidth]{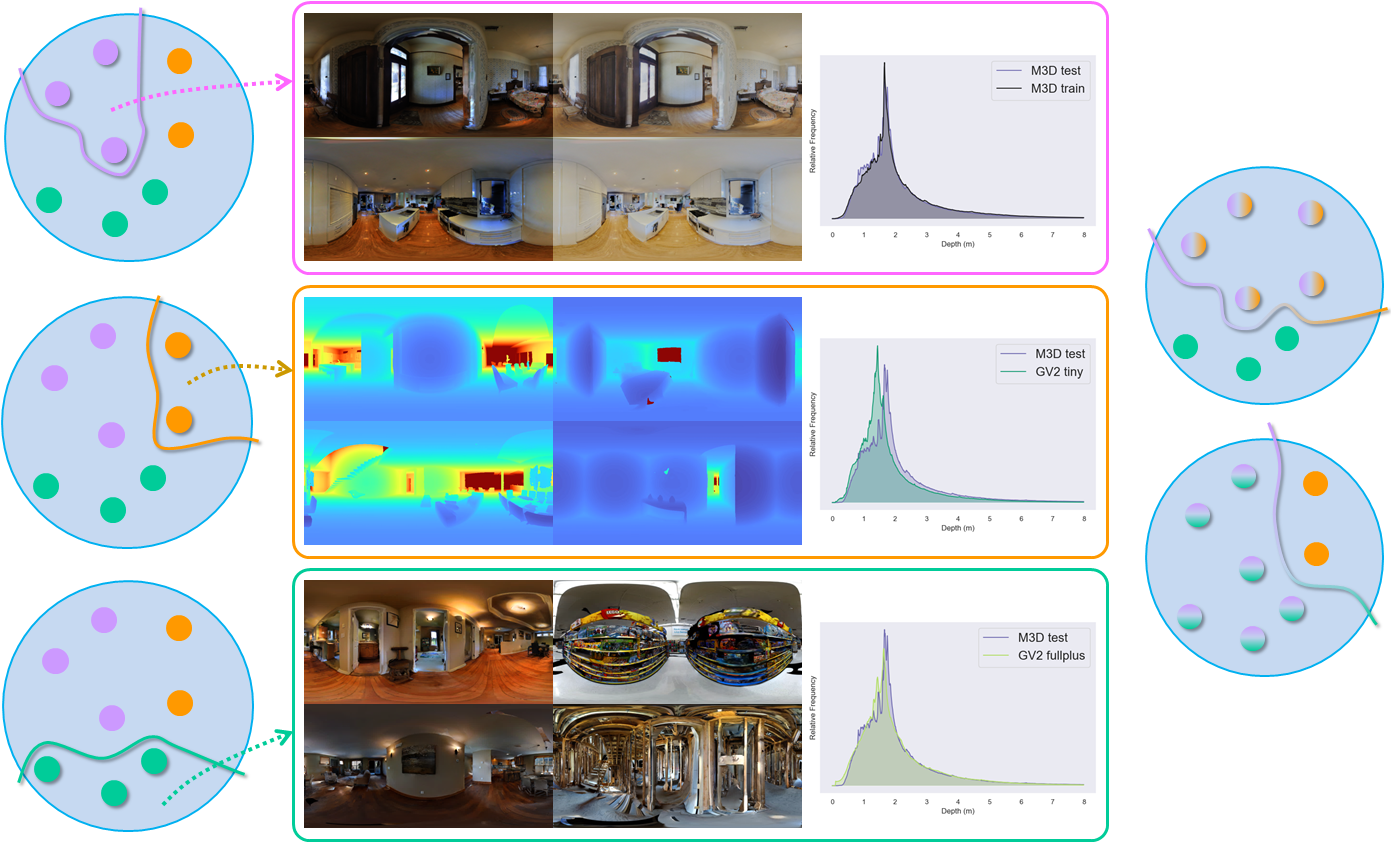}}
\caption{Disentangled distribution shifts. 
The left column represents the three singular shifts, namely \textcolor{drak_magenta}{covariate}, \textcolor{drak_orange}{prior} and \textcolor{drak_green}{concept}. 
In the middle, details regarding the composition and characteristics of the shifts can be found. 
The right column illustrates two stacked shift combinations.}
\label{fig:disentangled}
\end{figure*}

%% file: Figures/indicators_all.tex
\begin{figure*}[!htbp]
\centering
\subfloat{\includegraphics[width=0.33\linewidth]{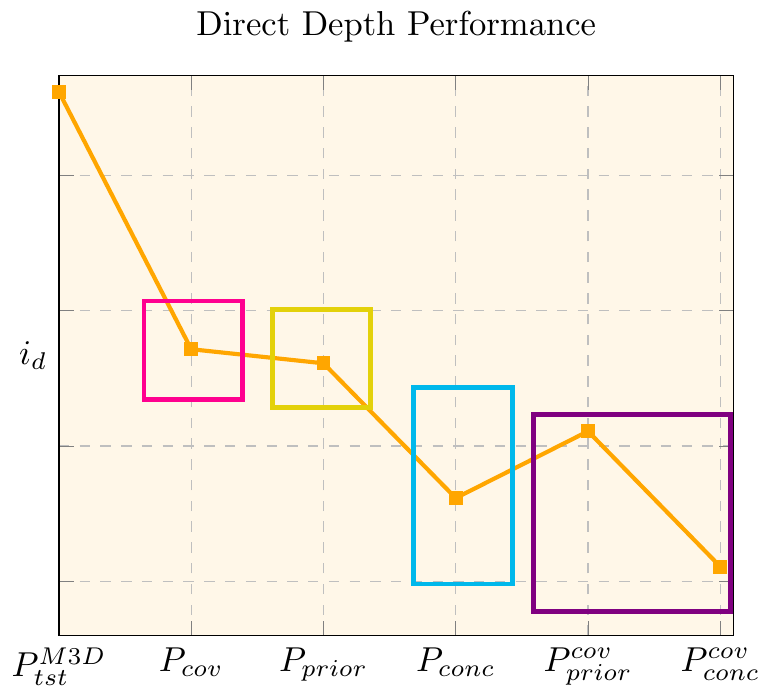}}
\subfloat{\includegraphics[width=0.33\linewidth]{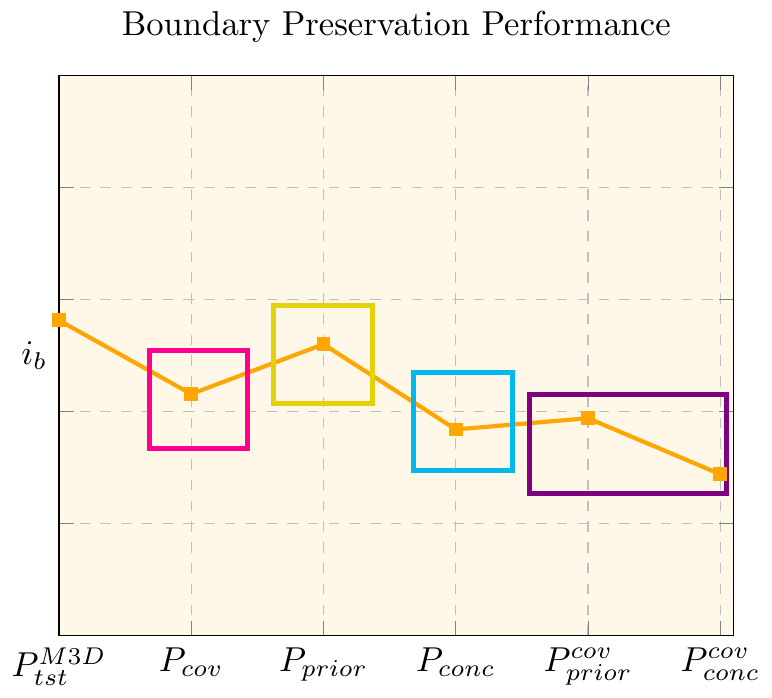}}
\subfloat{\includegraphics[width=0.33\linewidth]{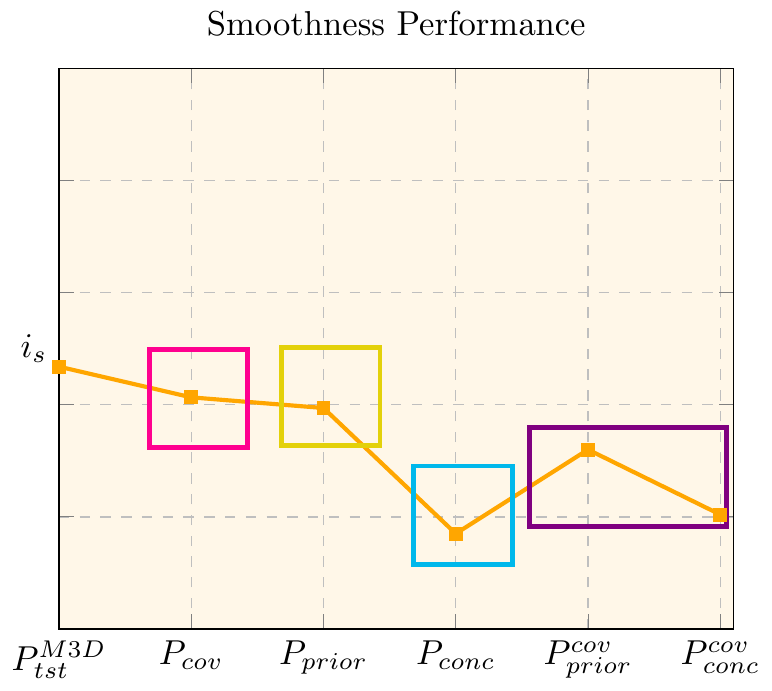}}
\caption{The effect of each distribution shift and their combinations.}
\label{fig:indicators_all}
\end{figure*}

%% file: Figures/indicators_augm.tex
\begin{figure*}[!htbp]
\centering
\subfloat{\includegraphics[width=0.33\linewidth]{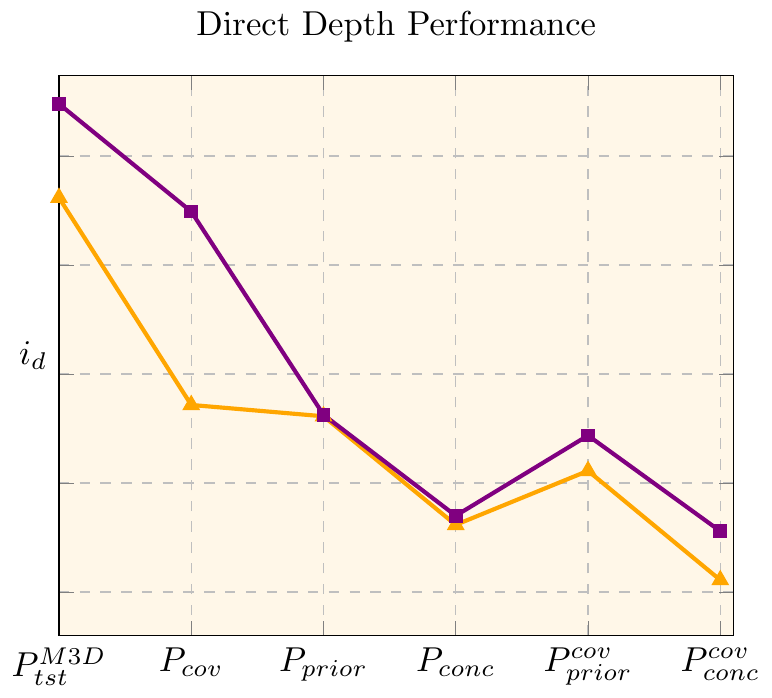}}
\subfloat{\includegraphics[width=0.33\linewidth]{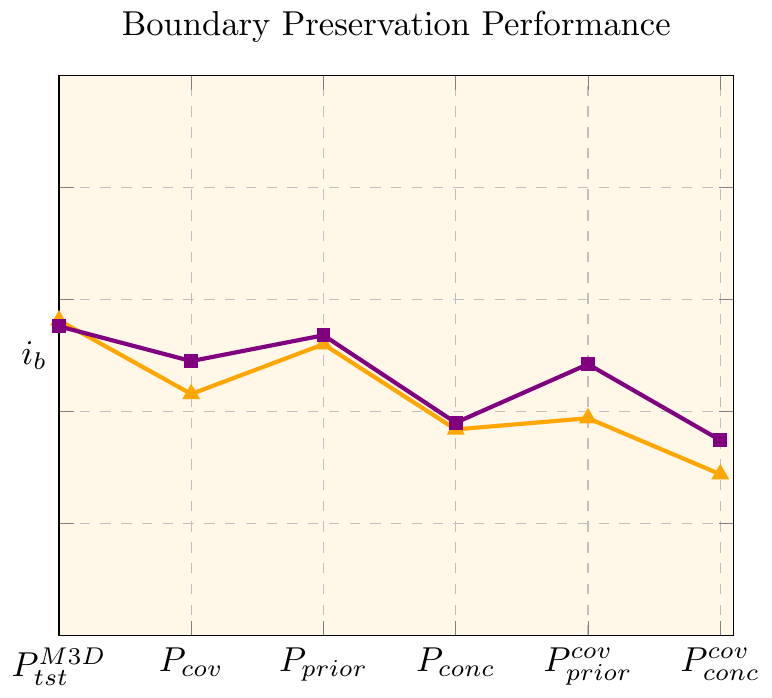}}
\subfloat{\includegraphics[width=0.33\linewidth]{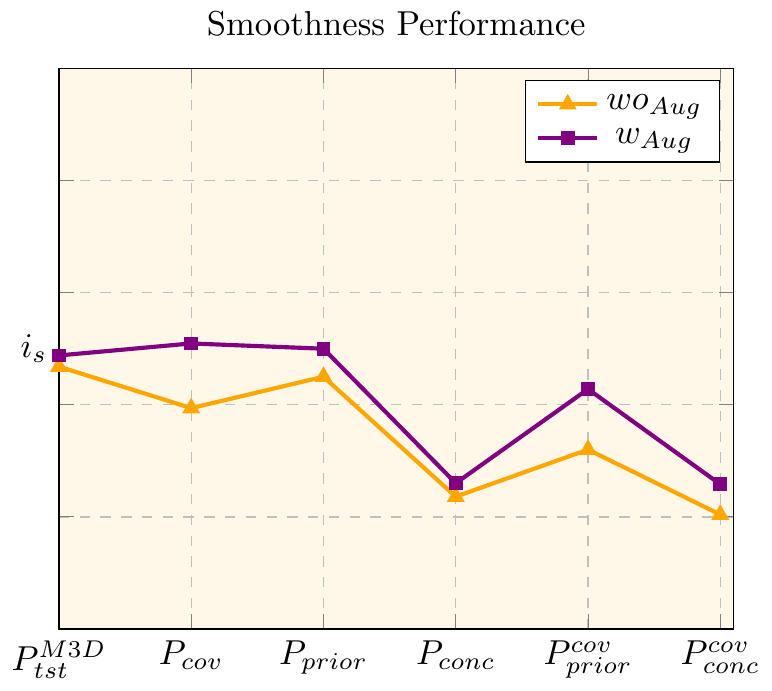}}
\caption{Photometric augmentations effect on the different distribution shifts.}
\label{fig:indicators_augm}

\end{figure*}

%% file: Figures/indicators_pnas.tex
\begin{figure*}[!htbp]
\centering
\subfloat{\includegraphics[width=0.33\linewidth]{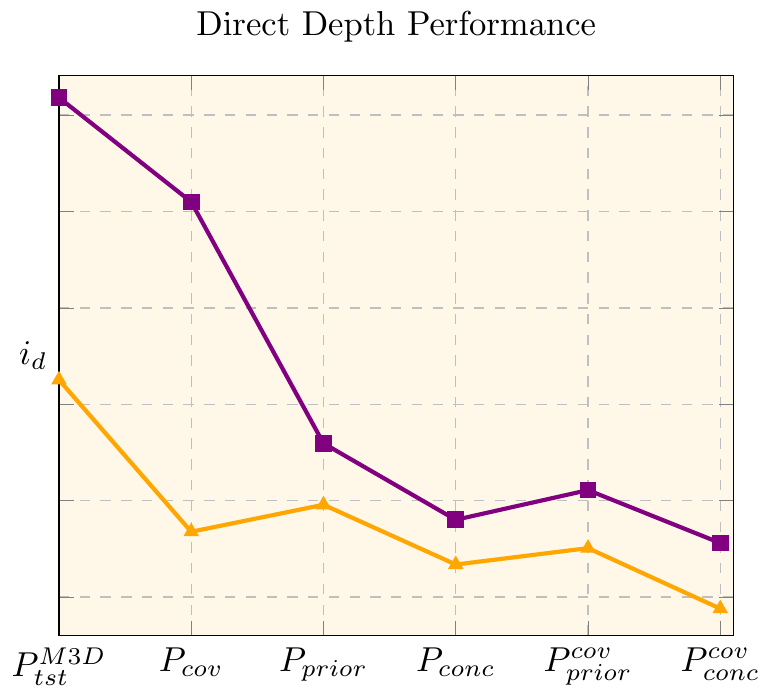}}
\subfloat{\includegraphics[width=0.33\linewidth]{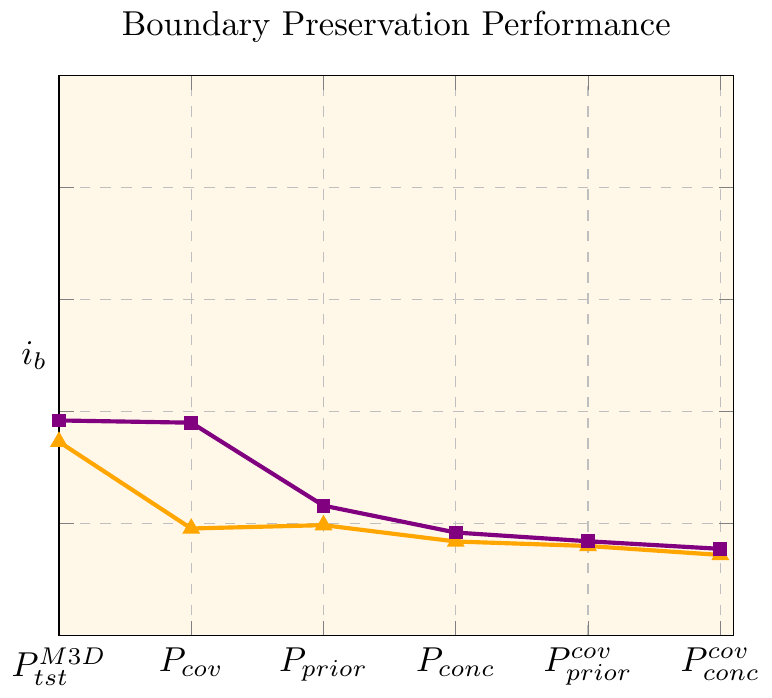}}
\subfloat{\includegraphics[width=0.33\linewidth]{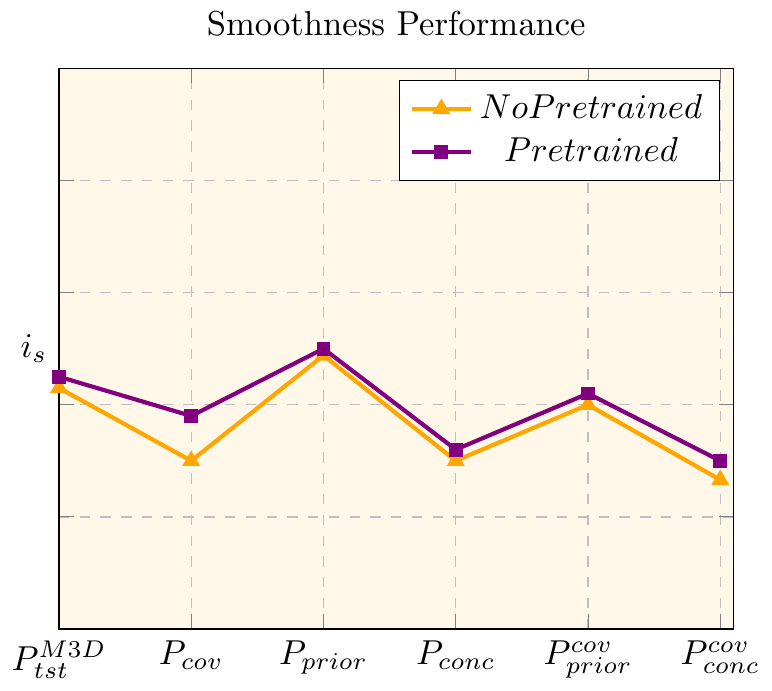}}
\caption{The effect of pretraining on our benchmark's different distribution shifts.}
\label{fig:indicators_pnas}
\end{figure*}

%% file: Tables/Table1_supp.tex
\begin{table*}[h]
\centering
\caption{Three-axis depth performance across models and data-splits. The \colorbox{drak_worst}{\textbf{worst}}, \colorbox{drak_second}{2\textsuperscript{nd} worst} and \colorbox{drak_third}{3\textsuperscript{rd} worst} performance drops per model are denoted in color respectively.}
\label{tab:Table1_supp}
\resizebox{\textwidth}{!}{%
\begin{tabular}{ll|ccccccccccccccccccccc}
\multirow{3}{*}{\rotatebox{90}{Model}}                   & \multicolumn{1}{c|}{\multirow{3}{*}{Split}}                      & \multicolumn{9}{c}{Direct Depth}                                                                                                                                                                                                                             & \multicolumn{8}{c}{Depth Discontinuity}                                                                                                                                                                                                                                                                                                   & \multicolumn{4}{c}{Depth Smoothness}                                                                              \\
                                                         & \multicolumn{1}{c|}{}                                            & \multicolumn{4}{c}{Error $\downarrow$}                                                             & \multicolumn{5}{c|}{Accuracy $\uparrow$}                                                                                                                & \multicolumn{2}{c}{Error $\downarrow$}                                                      & \multicolumn{6}{c|}{Accuracy $\uparrow$}                                                                                                                                                                                                    & Error $\downarrow$                                    & \multicolumn{3}{c}{Accuracy $\uparrow$}                   \\
                                                         & \multicolumn{1}{c|}{}                                            & \textit{$w$RMSE} & \textit{$w$RMSLE} & \textit{$w$AbsRel} & \multicolumn{1}{c|}{\textit{$w$SqRel}} & $\delta_{1.05}^{ico^6}$ & $\delta_{1.1}^{ico^6}$ & $\delta_{1.25}^{ico^6}$ & $\delta_{1.25^2}^{ico^6}$ & \multicolumn{1}{c|}{$\delta_{1.25^3}^{ico^6}$} & \textit{dbe}\textsuperscript{acc} & \multicolumn{1}{c|}{\textit{dbe}\textsuperscript{comp}} & \textit{prec}\textsubscript{$0.25$} & \textit{prec}\textsubscript{$0.5$} & \textit{prec}\textsubscript{$1$} & \textit{rec}\textsubscript{$0.25$} & \textit{rec}\textsubscript{$0.5$} & \multicolumn{1}{c|}{\textit{rec}\textsubscript{$1$}} & \multicolumn{1}{c|}{\textit{RMSE\textsuperscript{o}}} & $\alpha_{11.25^o}$ & $\alpha_{22.5^o}$ & $\alpha_{30^o}$  \\ \hline
\multirow{6}{*}{\rotatebox{90}{UNet}}                    & \textit{P\rlap{\textsuperscript{M3D}}\textsubscript{tst}}   & 0.452            & 0.130             & 0.115              & 0.081                                  & 36.68\%                 & 60.59\%                & 88.31\%                 & 96.96\%                   & \multicolumn{1}{c|}{98.73\%}                   & 1.270                             & 3.888                                                   & 58.97\%                             & \third{57.54\%}                    & \second{51.85\%}                 & 43.96\%                            & 36.69\%                           & \multicolumn{1}{c|}{28.59\%}                         & 16.021                                                & 61.80\%            & 76.58\%           & 81.70\%          \\
                                                         & \textit{P\textsubscript{cov}}                                    & \third{0.546}    & 0.130             & 0.135              & 0.113                                  & 29.08\%                 & 52.44\%                & 83.47\%                 & 83.68\%                   & \multicolumn{1}{c|}{95.28\%}                   & 1.526                             & 4.404                                                   & 63.64\%                             & 63.33\%                            & 57.23\%                          & \third{36.15\%}                    & \second{29.03\%}                  & \multicolumn{1}{c|}{\second{20.73\%}}                & 17.398                                                & 59.82\%            & 75.62\%           & 81.10\%          \\
                                                         & \textit{P\textsubscript{prior}}                                  & 0.472            & 0.206             & 0.173              & 0.141                                  & \second{21.85\%}        & \second{41.67\%}       & 81.49\%                 & 81.62\%                   & \multicolumn{1}{c|}{95.73\%}                   & 1.473                             & 4.338                                                   & 61.43\%                             & 64.51\%                            & 60.21\%                          & 46.53\%                            & 40.67\%                           & \multicolumn{1}{c|}{33.08\%}                         & 17.357                                                & 57.01\%            & 74.59\%           & 80.71\%          \\
                                                         & \textit{P\textsubscript{conc}}                                   & \second{0.617}   & \second{0.266}    & \third{0.184}      & \second{0.193}                         & 23.41\%                 & \third{42.42\%}        & \second{76.21\%}        & \second{76.44\%}          & \multicolumn{1}{c|}{\second{92.30\%}}          & \second{1.723}                    & \second{5.037}                                          & \second{54.45\%}                    & \second{56.37\%}                   & \third{52.31\%}                  & \second{34.61\%}                   & \third{29.07\%}                   & \multicolumn{1}{c|}{\third{23.02\%}}                 & \second{22.059}                                       & \second{46.84\%}   & \second{66.09\%}  & \second{73.41\%} \\
                                                         & \textit{P\rlap{\textsuperscript{cov}}\textsubscript{prior}} & 0.545            & \third{0.232}     & \second{0.185}     & \third{0.185}                          & \third{22.82\%}         & 42.82\%                & \third{79.43\%}         & \third{79.58\%}           & \multicolumn{1}{c|}{\third{93.73\%}}           & \third{1.694}                     & \third{4.844}                                           & \third{57.63\%}                     & 59.49\%                            & 53.19\%                          & 37.47\%                            & 31.28\%                           & \multicolumn{1}{c|}{23.22\%}                         & \third{19.219}                                        & \third{53.24\%}    & \third{71.44\%}   & \third{78.00\%}  \\
                                                         & \textit{P\rlap{\textsuperscript{cov}}\textsubscript{conc}}  & \worst{0.737}    & \worst{0.297}     & \worst{0.220}      & \worst{0.411}                          & \worst{20.96\%}         & \worst{38.47\%}        & \worst{70.20\%}         & \worst{70.46\%}           & \multicolumn{1}{c|}{\worst{87.99\%}}           & \worst{1.948}                     & \worst{5.560}                                           & \worst{50.65\%}                     & \worst{50.90\%}                    & \worst{44.46\%}                  & \worst{26.76\%}                    & \worst{21.28\%}                   & \multicolumn{1}{c|}{\worst{15.46\%}}                 & \worst{23.898}                                        & \worst{43.54\%}    & \worst{63.06\%}   & \worst{70.69\%}  \\ \hline
\multirow{6}{*}{\rotatebox{90}{UNet\textsubscript{aug}}} & \textit{P\rlap{\textsuperscript{M3D}}\textsubscript{tst}}   & 0.433            & 0.068             & 0.109              & 0.073                                  & 37.36\%                 & 63.11\%                & 89.59\%                 & 89.76\%                   & \multicolumn{1}{c|}{97.42\%}                   & 1.360                             & 3.876                                                   & 64.82\%                             & 64.94\%                            & 60.41\%                          & 44.96\%                            & 37.02\%                           & \multicolumn{1}{c|}{27.96\%}                         & 15.099                                                & 63.99\%            & 77.98\%           & 82.83\%          \\
                                                         & \textit{P\textsubscript{cov}}                                    & 0.469            & 0.073             & 0.117              & 0.091                                  & 35.35\%                 & 61.31\%                & 88.20\%                 & 88.40\%                   & \multicolumn{1}{c|}{96.83\%}                   & 1.443                             & 4.156                                                   & 64.27\%                             & 63.79\%                            & \third{58.79\%}                  & \third{42.17\%}                    & \third{34.00\%}                   & \multicolumn{1}{c|}{\third{25.21\%}}                 & 15.653                                                & 63.92\%            & 78.34\%           & 83.34\%          \\
                                                         & \textit{P\textsubscript{prior}}                                  & 0.458            & 0.084             & 0.170              & 0.102                                  & \second{20.43\%}        & \second{39.73\%}       & 81.19\%                 & 81.32\%                   & \multicolumn{1}{c|}{96.19\%}                   & 1.448                             & 4.268                                                   & 62.69\%                             & 66.19\%                            & 62.27\%                          & 47.56\%                            & 41.51\%                           & \multicolumn{1}{c|}{32.90\%}                         & 16.307                                                & 59.48\%            & 76.41\%           & 82.16\%          \\
                                                         & \textit{P\textsubscript{conc}}                                   & \second{0.601}   & \second{0.103}    & \second{0.176}     & \second{0.152}                         & 23.61\%                 & 42.70\%                & \second{76.98\%}        & \second{77.22\%}          & \multicolumn{1}{c|}{\second{92.78\%}}          & \second{1.704}                    & \second{5.006}                                          & \second{56.24\%}                    & \second{58.18\%}                   & \second{53.33\%}                 & \second{35.45\%}                   & \second{29.78\%}                  & \multicolumn{1}{c|}{\worst{23.07\%}}                 & \second{20.870}                                       & \second{49.29\%}   & \second{68.06\%}  & \second{75.09\%} \\
                                                         & \textit{P\rlap{\textsuperscript{cov}}\textsubscript{prior}} & \third{0.475}    & \third{0.087}     & \third{0.174}      & \third{0.114}                          & \worst{20.22\%}         & \worst{39.51\%}        & \third{80.39\%}         & \third{80.52\%}           & \multicolumn{1}{c|}{\third{95.70\%}}           & \third{1.533}                     & \third{4.392}                                           & \third{60.69\%}                     & \third{63.32\%}                    & 59.43\%                          & 44.54\%                            & 38.01\%                           & \multicolumn{1}{c|}{29.63\%}                         & \third{16.669}                                        & \third{58.74\%}    & \third{75.78\%}   & \third{81.62\%}  \\
                                                         & \textit{P\rlap{\textsuperscript{cov}}\textsubscript{conc}}  & \worst{0.624}    & \worst{0.108}     & \worst{0.183}      & \worst{0.170}                          & \third{22.78\%}         & \third{41.57\%}        & \worst{75.56\%}         & \worst{75.80\%}           & \multicolumn{1}{c|}{\worst{92.06\%}}           & \worst{1.769}                     & \worst{5.148}                                           & \worst{54.64\%}                     & \worst{55.56\%}                    & \worst{50.02\%}                  & \worst{32.85\%}                    & \worst{26.91\%}                   & \multicolumn{1}{c|}{\worst{20.50\%}}                 & \worst{21.234}                                        & \worst{48.58\%}    & \worst{67.49\%}   & \worst{74.58\%}  \\ \hline
\multirow{6}{*}{\rotatebox{90}{Pnas}}                    & \textit{P\rlap{\textsuperscript{M3D}}\textsubscript{tst}}   & 0.561            & 0.085             & 0.133              & 0.120                                  & 32.69\%                 & 56.94\%                & 96.30\%                 & 95.38\%                   & \multicolumn{1}{c|}{97.95\%}                   & 2.654                             & 5.730                                                   & 38.73\%                             & 30.26\%                            & 23.58\%                          & 18.74\%                            & 10.48\%                           & \multicolumn{1}{c|}{8.48\%}                          & 20.118                                                & 53.88\%            & 69.81\%           & 75.65\%          \\
                                                         & \textit{P\textsubscript{cov}}                                    & \second{0.703}   & 0.109             & 0.160              & 0.159                                  & 23.45\%                 & 45.12\%                & 76.27\%                 & 77.79\%                   & \multicolumn{1}{c|}{92.06\%}                   & 2.869                             & 6.075                                                   & 36.70\%                             & 28.00\%                            & \third{18.99\%}                  & \third{12.33\%}                    & \second{6.80\%}                   & \multicolumn{1}{c|}{\second{5.32\%}}                 & 21.486                                                & 52.07\%            & 68.75\%           & 74.91\%          \\
                                                         & \textit{P\textsubscript{prior}}                                  & 0.562            & 0.098             & 0.188              & 0.146                                  & \second{19.67\%}        & \second{38.39\%}       & 76.37\%                 & 77.53\%                   & \multicolumn{1}{c|}{94.28\%}                   & 2.651                             & 5.243                                                   & 34.12\%                             & 29.20\%                            & 23.15\%                          & 18.43\%                            & 11.70\%                           & \multicolumn{1}{c|}{9.68\%}                          & 19.929                                                & 52.64\%            & 70.83\%           & 77.51\%          \\
                                                         & \textit{P\textsubscript{conc}}                                   & \third{0.693}    & \second{0.117}    & \second{0.200}     & \second{0.196}                         & 21.27\%                 & \third{39.37\%}        & \second{72.84\%}        & \second{73.10\%}          & \multicolumn{1}{c|}{\second{90.80\%}}          & \third{3.192}                     & \second{7.277}                                          & \third{32.20\%}                     & \third{25.87\%}                    & 19.51\%                          & 13.69\%                            & 8.32\%                            & \multicolumn{1}{c|}{6.68\%}                          & \second{24.433}                                       & \second{44.01\%}   & \second{63.45\%}  & \second{71.07\%} \\
                                                         & \textit{P\rlap{\textsuperscript{cov}}\textsubscript{prior}} & 0.663            & \third{0.116}     & \third{0.192}      & \third{0.170}                          & \third{21.21\%}         & 40.24\%                & \third{74.54\%}         & \third{74.72\%}           & \multicolumn{1}{c|}{\third{91.12\%}}           & \second{3.266}                    & \third{7.251}                                           & \second{29.89\%}                    & \second{24.68\%}                   & \second{17.70\%}                 & \second{11.58\%}                   & \third{7.01\%}                    & \multicolumn{1}{c|}{\third{5.58\%}}                  & \third{22.493}                                        & \third{47.89\%}    & \third{66.76\%}   & \third{73.90\%}  \\
                                                         & \textit{P\rlap{\textsuperscript{cov}}\textsubscript{conc}}  & \worst{0.842}    & \worst{0.145}     & \worst{0.222}      & \worst{0.244}                          & \worst{18.50\%}         & \worst{34.75\%}        & \worst{65.66\%}         & \worst{65.93\%}           & \multicolumn{1}{c|}{\worst{85.70\%}}           & \worst{3.674}                     & \worst{7.881}                                           & \worst{28.13\%}                     & \worst{21.04\%}                    & \worst{13.40\%}                  & \worst{8.15\%}                     & \worst{4.65\%}                    & \multicolumn{1}{c|}{\worst{3.70\%}}                  & \worst{26.583}                                        & \worst{39.92\%}    & \worst{59.69\%}   & \worst{67.72\%}  \\ \hline
\multirow{6}{*}{\rotatebox{90}{Pnas\textsubscript{pre}}} & \textit{P\rlap{\textsuperscript{M3D}}\textsubscript{tst}}   & 0.467            & 0.070             & 0.107              & 0.086                                  & 40.90\%                 & 64.98\%                & 90.38\%                 & 90.56\%                   & \multicolumn{1}{c|}{97.33\%}                   & 2.217                             & 5.019                                                   & 44.35\%                             & 37.55\%                            & 31.57\%                          & 25.78\%                            & 15.54\%                           & \multicolumn{1}{c|}{11.60\%}                         & 17.785                                                & 59.34\%            & \third{73.58\%}   & \third{78.80\%}  \\
                                                         & \textit{P\textsubscript{cov}}                                    & 0.492            & 0.074             & 0.114              & 0.094                                  & 39.53\%                 & 62.86\%                & 88.92\%                 & 89.14\%                   & \multicolumn{1}{c|}{96.92\%}                   & 2.304                             & \third{6.118}                                           & 44.83\%                             & 37.46\%                            & \third{31.06\%}                  & \third{24.20\%}                    & \third{14.55\%}                   & \multicolumn{1}{c|}{\third{10.87\%}}                 & \third{18.066}                                        & 60.24\%            & 74.64\%           & 79.93\%          \\
                                                         & \textit{P\textsubscript{prior}}                                  & 0.501            & 0.087             & 0.172              & 0.112                                  & \second{18.89\%}        & \second{37.30\%}       & 80.83\%                 & 80.99\%                   & \multicolumn{1}{c|}{96.34\%}                   & 2.307                             & 5.936                                                   & 40.62\%                             & 37.67\%                            & 34.14\%                          & 26.48\%                            & 18.37\%                           & \multicolumn{1}{c|}{14.72\%}                         & 18.003                                                & \third{57.90\%}    & 74.54\%           & 80.52\%          \\
                                                         & \textit{P\textsubscript{conc}}                                   & \second{0.616}   & \second{0.103}    & \third{0.174}      & \second{0.149}                         & 23.20\%                 & 41.88\%                & \second{77.59\%}        & \second{77.87\%}          & \multicolumn{1}{c|}{\second{93.24\%}}          & \second{2.658}                    & \second{6.712}                                          & \second{38.10\%}                    & \second{32.81\%}                   & \second{27.60\%}                 & \second{20.23\%}                   & \second{13.14\%}                  & \multicolumn{1}{c|}{\second{10.46\%}}                & \worst{22.060}                                        & \worst{49.32\%}    & \worst{67.63\%}   & \worst{74.63\%}  \\
                                                         & \textit{P\rlap{\textsuperscript{cov}}\textsubscript{prior}} & \third{0.531}    & \third{0.093}     & \worst{0.189}      & \third{0.130}                          & \worst{16.68\%}         & \worst{32.96\%}        & \third{78.06\%}         & \third{78.21\%}           & \multicolumn{1}{c|}{\third{95.67\%}}           & \third{2.400}                     & 6.094                                                   & \third{39.70\%}                     & \third{36.16\%}                    & 32.07\%                          & 24.74\%                            & 16.74\%                           & \multicolumn{1}{c|}{13.23\%}                         & 17.949                                                & 58.22\%            & 74.70\%           & 80.59\%          \\
                                                         & \textit{P\rlap{\textsuperscript{cov}}\textsubscript{conc}}  & \worst{0.649}    & \worst{0.109}     & \second{0.184}     & \worst{0.161}                          & \third{21.98\%}         & \third{39.44\%}        & \worst{75.48\%}         & \worst{75.77\%}           & \multicolumn{1}{c|}{\worst{92.34\%}}           & \worst{2.790}                     & \worst{6.918}                                           & \worst{37.29\%}                     & \worst{31.64\%}                    & \worst{25.48\%}                  & \worst{18.25\%}                    & \worst{11.48\%}                   & \multicolumn{1}{c|}{\worst{9.10\%}}                  & \second{22.019}                                       & \second{49.63\%}   & \second{67.78\%}  & \second{74.69\%} \\ \hline
\end{tabular}%
}
\end{table*}